# The Future of Human-AI Collaboration: A Taxonomy of Design Knowledge for Hybrid Intelligence Systems

Dominik Dellermann
vencortex
dominik.dellermann@vencortex.com

Adrian Calma
vencortex
adrian.calma@vencortex.com

Nikolaus Lipusch
Kassel University
lipusch@uni-kassel.de

Thorsten Weber
Kassel University
weber@uni-kassel.de

Sascha Weigel
Kassel University
weigel@uni-kassel.de

Philipp Ebel
University of St. Gallen
ebel@unisg.ch

## Abstract

*Recent technological advances, especially in the field of machine learning, provide astonishing progress on the road towards artificial general intelligence. However, tasks in current real-world business applications cannot yet be solved by machines alone. We, therefore, identify the need for developing socio-technological ensembles of humans and machines. Such systems possess the ability to accomplish complex goals by combining human and artificial intelligence to collectively achieve superior results and continuously improve by learning from each other. Thus, the need for structured design knowledge for those systems arises. Following a taxonomy development method, this article provides three main contributions: First, we present a structured overview of interdisciplinary research on the role of humans in the machine learning pipeline. Second, we envision hybrid intelligence systems and conceptualize the relevant dimensions for system design for the first time. Finally, we offer useful guidance for system developers during the implementation of such applications.*

## 1. Introduction

Recent technological advances especially in the field of deep learning provide astonishing progress on the road towards artificial general intelligence (AGI) [1, 2]. Artificial intelligence (AI) are progressively achieving (super-) human level performance in various tasks such as autonomous driving [3], cancer detection [4], or playing complex games [5, 6]. Therefore, more and more business applications that are based on AI technologies arise. Both research and practice are wondering when AI will be able to solve complex tasks in real-world business applications apart from laboratory settings in research. However, those advances provide a rather one-sided picture on AI, denying the fact that although AI is capable to solve certain tasks with quite impressive performance, AGI is far away from being achieved. There are lots of problems that machines can not yet solve alone [7], such as applying expertise to decision making, planning, or creative tasks just to name a few. In particular, machine learning systems in the wild have major difficulties with being adaptive to dynamic environments and self adjusting [8], lack of what humans call common sense. This makes them highly vulnerable for adversarial examples [9]. Moreover, AGI needs massive amounts of training data compared to humans, who can learn from only few examples [10], and fails to work with certain data types (e.g. soft data). Nevertheless, a lack of control of the learning process might lead to unintended consequences (e.g. racism biases) and limit interpretability, which is crucial for critical domains such as medicine [11]. Therefore, humans are still required at various positions in the loop of the machine learning process. While lot of work has been done in creating training sets with human labelers, more recent research point towards end user involvement [12] and teaching of such machines [5], thus, combining humans and machines in **hybrid intelligence** systems. The main idea of hybrid intelligence systems is, thus, that socio-technical ensembles and its human and AI parts can co-evolve to improve over time. Therefore, the following central questions are arise: *Which and how should certain design decisions be made for implementing such systems?* The purpose of this paper is to point towards such hybrid intelligence systems. Thereby, we aim at conceptualizing the idea of **hybrid intelligence** systems and provide an initial taxonomy of design knowledge for developing such socio-technical ensembles. By following a taxonomy development method [13], we reviewed various literature in interdisciplinary fields and combine those findings with empirical examination of practical business applications in the context of

*hybrid intelligence*. The contribution of this paper is threefold. First, we provide a structured overview of interdisciplinary research on the role of humans in the machine learning pipeline. Second, we offer an initial conceptualization of the term *hybrid intelligence* systems and relevant dimensions for system design. Third, we intend to provide useful guidance for system developers during the implementation of hybrid intelligence systems in real-world applications. Towards this end, we propose an initial taxonomy of hybrid intelligence systems.

## 2. Related Work

### 2.1. Machine Learning and AI

The subfield of intelligence that relates to machines is called **artificial intelligence** (**AI**). By this term we mean systems that perform *"[] activities that we associate with human thinking, activities such as decision-making, problem solving, learning []"* [14]. Although, various definitions exist for AI, this term generally covers the idea of creating machines that can accomplish complex goals. This includes facets such as natural language processing, perceiving objects, storing of knowledge and applying it for solving problems, and machine learning to adapt to new circumstances and act in its environment [15].

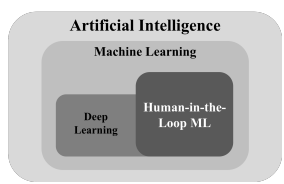

**Figure 1. Machine learning and AI.**

A subset of techniques that is required to achieve AI is **machine learning (ML)**. Mitchell [16] defines this as: *"A computer program is said to learn from experience E with respect to some class of tasks T and performance measure P, if its performance at tasks in T, as measured by P, improves with experience E."*

A popular approach that drives current progress in both paradigms is **deep learning** [9]. Deep-learning constitutes a representation-learning method that includes multiple levels of representation, obtained by combining simpler but non-linear models. Each of those models transforms the representation of one level (starting with the input data) into a representation at more abstract level [17]. Deep learning is a special machine learning technique.

Finally, **human-in-the-loop learning** describes machine learning approaches (both deep and other) that use the human in some part of the pipeline. Such approaches are in contrast to research on most knowledge-base systems in IS that use rather static knowledge repositories. We will focus on this in the following chapter.

### 2.2. The Role of Humans-in-the-Loop of Machine Learning

Although, the terms of AI and machine learning give the impression that humans become to some extent obsolete, the machine learning pipeline still requires lot of human interaction such as for feature engineering, parameter tuning, or training. While deep learning has decreased the effort for manual feature engineering and some automation approaches (e.g. AutoML [18]) support human experts in tuning models, the human is still heavily in the loop for sense-making and training. For instance, unsupervised learning requires humans to make sense of clusters that are identified as patterns in data to create knowledge [19]. More obviously, human input is required to train models in supervised machine learning approaches, especially for creating training data, debug models, or train algorithms such as in reinforcement learning [5]. This is especially relevant when divergences of real-life and machine learning problem formulations emerge. This is for instance the case when static (offline) training datasets are not perfectly representative of realist and dynamic environments [20]. Moreover, human input is crucial when models need to learn from human preferences (e.g. recommender systems) and adapt to users or when security concerns require both control and interpretability of the learning process and the output [11]. Therefore, more recent research has focused on interactive forms of learning (e.g. [21, 12] and machine teaching (e.g. [22]). Those approaches make active use of human input (e.g. active learning [23]) and thus learn from human intelligence. This allows machines to learn tasks that they can not yet achieve alone [7], adapt to environmental dynamics, and deal with unknown situations [24].

### 2.3. Hybrid Intelligence

Rather then using the human just in certain parts and time during the process of creating machine learning models, applications that are able to deal with real-world problems require a continuously collaborating socio-technological ensemble integrating humans and machines, which is contrast to previous research on decision support and expert systems [21, 25].

Therefore, we argue that the most likely paradigm for the division of labor between humans and machines in the next years, or probably decades, is **hybrid intelligence**. This concept aims at using the complementary strengths of human intelligence and AI to behave more intelligently than each of the two could be in separation (e.g. [7]). The basic rational is to try to combine the complementary strengths of heterogeneous intelligences (i.e., human and artificial agents) into a socio-technological ensemble. We envision **hybrid intelligence** systems, which are defined as systems that have *the ability to accomplish complex goals by combining human and artificial intelligence to collectively achieve superior results than each of the could have done in separation and continuously improve by learning from each other.*

*Collectively:* means that tasks are performed collectively. This means that the activities conducted by each part are dependent, however, are not necessarily always aligned to achieve a common goal (e.g. teaching an AI adversarial tasks such as playing games).

*Superior results:* defines that the system achieves a performance that none of the involved actors could have achieved in a certain task without the other. The goal is, therefore, to make the outcome (e.g. a prediction) both more efficient and effective on the level of the whole socio-technical system by achieving goals that could not have been achieved before.

*Continuous learning:* describes that over time this socio-technological system improves both as a whole and each single component (i.e. humans and machines) learn through experience from each other, thus improving performance in a certain task. The performance of such systems can be thus not only measured by the superior outcome of the whole system but also by the learning of the human and machine agents that are parts of the socio-technical system.

The idea of hybrid intelligence systems is thus that socio-technical ensembles and its human and AI parts can co-evolve to improve over time. The central questions are, therefore, which and how certain design decisions should be made for implementing such hybrid systems rather than focusing.

## 3. Methodology

### 3.1. Taxonomy Development Method

For developing our proposed taxonomy, we followed the methodological procedures of Nickerson et al. [13]. In general, a taxonomy is defined as a *"fundamental mechanism for organizing knowledge"* and the term is considered as a synonym to *"classification"* and *"typology"* [13]. The method follows an iterative process consisting of the following steps: 1) defining a meta-characteristic; 2) determining stopping conditions; 3) selecting an empirical-to-conceptual or conceptual-to-empirical approach; and 4) iteratively following this approach, until the stopping conditions are met (see Figure 3).

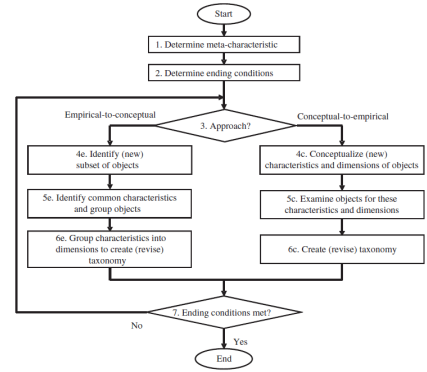

**Figure 2.** Taxonomy development method [13].

The process of the taxonomy development starts with defining a set of meta-characteristic. This step limits the odds of *naive empiricism* where a large number of characteristics are defined in search for random pattern, and reflects the expected application of the taxonomy [13]. For this purpose, we define those meta-characteristic as generic design dimensions that are required for developing hybrid intelligence systems. Based on our classification from literature, we choose four dimensions: task characteristics, learning paradigm, human-AI interaction, and AI-human interaction. In the second step, we selected both objective and subjective conditions to conclude the iterative process. The following conditions, adapted from Nickerson et al. [13], were selected:

We applied the following **objective conditions**: 1) All papers from the sample of the literature review and empirical cases are examined. 2) Then, at least one object is classified under every characteristic of every dimension. 3) While performing the last iteration, no new dimension or characteristics are added. 4) We treated every dimension as unique. 5) Lastly, every characteristic is unique within its dimension.

The following **subjective conditions** were considered: conciseness, robustness, comprehensiveness, extensibility, explanatory, and information availability. We included no unnecessary dimension or characteristic (conciseness), whereas there are enough dimensions and characteristics to differentiate (robustness). At this point, all design decisions can be classified in the taxonomy (comprehensiveness), while still allowing for new dimensions and characteristics to be subsequently added (extensible). Furthermore, the information is valuable for guiding hybrid intelligence systems design decisions (explanatory) and is typically available or easily interpretable (information availability).

**Table 1. Empirical evidence from business applications of hybrid intelligence systems.**

| Application | Domain | Reference |
|---|---|---|
| Teachable Machine | Image Recognition | [26] |
| Cindicator | Asset Management | [27] |
| vencortex | Startup Financing | [28] |
| Cobi | Conference Sheduling | [29] |
| Stitch Fix | Fashion | [30] |
| Alpha Go | Games | [31] |
| Custom Decision Service | General | [32] |
| TOTAL | 7 | |

We conducted a total of three iterations so far. The first iteration used a conceptual-to-empirical approach, where we used extant theoretical knowledge from literature in various fields such as computer science, HCI, information systems, and neuro science to guide the initial dimensions and characteristics of the taxonomy. Based on the identified dimensions of *hybrid intelligence* systems, we sampled seven real-world applications that make use of human and AI combinations. The second iteration used the empirical-to conceptual approach focuses on creating characteristics and dimensions based on the identification of common characteristics from a sample of AI applications in practice. The third iteration then used the conceptual-to-empirical approach, based on an extended literature review including newly identified search termini.

### 3.2. Data Sources and Sample

**Literature review:** For conducting our literature review, we followed the methodological procedures of [33, 34]. The literature search was conducted from April to June 2018. A prior informal literature search revealed keywords for the database searches resulting in the search string ("hybrid intelligence" OR "human-in-the-loop" OR "interactive machine learning" OR "machine teaching" OR "machine learning AND crowdsourcing" OR "human supervision" OR "human understandable machine learning" OR "human concept learning"). During this initial phase we decided to exclude research on knowledge-base systems such as expert systems or decision support systems in IS [35, 25], as the studies either do not focus on the continuous learning of the knowledge repository or do not use machine learning techniques at all. Moreover, the purpose of this study is to identify and classify relevant (socio-) technical design knowledge for hybrid intelligence systems, which is also not included in those studies. The database search was constrained to title, abstract, keywords and not limited to a certain publication. Databases include AISeL, IEEE Xplore, ACM DL, AAAI DL, and arXiv to identify relevant interdisciplinary literature from the fields of IS, HCI, bio-informatics, and computer science. The search resulted in a total of 2505 hits. Titles, abstracts and keywords were screened for potential fit to the purpose of our study. Screening was conducted by three researchers independently and resulted in 85 articles that were reviewed in detail so far. A backward and forward search ensured the extensiveness of our results. Table 1 lists the number of search results after the review phases.

**Empirical cases:** To extend our findings from literature and provide empirical evidence (cf. Table 1) from recent (business) applications of *hybrid intelligence* systems, we include an initial set of seven empirical applications that was analyzed for enhancing our taxonomy.

## 4. Taxonomy of Design Knowledge on hybrid intelligence Systems

Our taxonomy of hybrid intelligence systems is organized along the four meta-dimensions *task characteristics, learning paradigm, human-AI interaction*, and *AI-human interaction*. Moreover, we identified 16 sub-dimensions and a total of 50 categories for the proposed taxonomy. For organizing the dimensions of the taxonomy we followed a hierarchical approach following the sequence of the design decisions that are necessary to develop such systems.

### 4.1. Task Characteristics

The goal of hybrid intelligence is to create superior results through a collaboration between humans and machines. The central component that drives design decisions for *hybrid intelligence* systems is the task, that humans and machines solve collaboratively. Task characteristics focus on how the task itself is carried out [36]. In context of *hybrid intelligence* systems, we identify the following four important tasks characteristics.

**Type of Task:** The task to be solved is the first dimension that has to be defined for developing *hybrid intelligence* systems. In this context, we identified four generic categories of tasks: *recognition, prediction, reasoning* and *action*. First, *recognition* defines tasks that recognize for instance objects [17], images [37], or natural language [38]. On an application level such tasks are used for autonomous driving (e.g. [3]) or smart assistants such as Alexa, Siri or Duplex. Second, *prediction* tasks aim at predicting future events based on previous data such as stock prices or market dynamics [39]. The third type of task, *reasoning*, focuses on understanding data by for instance inductively building (mental) models of a certain phenomenon and therefore make it possible to solve complex problems with small amount of data [10]. Finally, *action* tasks are characterized as such that require an agent (human or machine) to conduct a certain kind of action [40].

**Goals:** The two involved agents, the human and the AI, may have a *common* "goal" like solving a problem through the combination of the knowledge and abilities of both. An example for such common goals are recommender systems (e.g. Netflix [41]), which learn a user's decision model to offer suggestions. In other contexts, the agents goals also be *adversarial*. For instance, in settings where AIs tzry to beat human in games such as IBMs Watson in the game of Jeopardy! [42]. In many other cases the goal of the human and the AI may also be *independent* for example when humans train image classifiers without being involved in the end solution.

**Shared Data Representation:** The shared data representation is what is the data that is shown to both the human and the machine before executing their tasks. The data can be represented in different levels of granularity and abstraction to create a shared understanding between humans and machines [22, 43]. *Features* describe phenomenas in different kinds of dimensions like height and weight of a human being. *Instances* are examples of a phenomena which are specified by features. *Concepts* on the other hand are multiple instances that belong to one common theme, e.g. pictures of different humans. *Schemas* finally illustrate relations between different concepts [44].

**Timing in Machine Learning Pipeline:** The last sub-dimension describes the timing in the machine learning pipeline that focuses on hybrid intelligence. For this dimension we identified three characteristics: *feature engineering*, *parameter tuning*, and *training*. First, *feature engineering* allows the integration of domain knowledge in machine learning models. While more recent advances make it possible to fully automatically (i.e. machine only) learn features through deep learning, human input can be combined for creating and enlarging features such in the case of artist identification on images and quality classification of Wikipedia articles (e.g. [45]). Second, *parameter tuning* is applied to optimize models. Here machine learning experts typically use their deep understanding of statistical models to tune hyper-parameters or select models. Such human only parameter tuning can be augmented with approaches such as AutoML [18] or neural architecture search [46, 47] automate the design of machine learning models,thus, making it much more accessible for non-experts. Finally, human input is crucial for *training* machine learning models in many domains. For instance large dataset such as ImageNet or the lung cancer dataset LUNA16 rely on human annotations. Moreover, recommender systems heavily rely on input of human usage behavior to adapt to specific preferences (e.g. [12]) and robotic applications are trained by human examples [40].

Figure 3. Taxonomy of hybrid intelligence design.

## 4.2. Learning Paradigm

**Augmentation:** In general, *hybrid intelligence* systems allow three different forms of augmentation: *human*, *machine*, and *hybrid* augmentation. The augmentation of *human* intelligence is focused on typically applications that enable humans to solve tasks through the predictions of an algorithm such as in financial forecasting or solving complex problems [48]. Contrary, most research in the field of machine learning focuses on leveraging human input for training to augment *machines* for solving tasks that they cannot yet solve alone [7]. Finally, more recent work identified the great potential for simultaneously augmenting both at the same time through *hybrid* augmentation [49, 50] or the example of Alpha Go that started by learning from human game moves (i.e. *machine* augmentation) and finally offered hybrid augmentation by inventing creative moves that taught even mature players novel strategies [6, 51].

**Machine Learning Paradigm:** The machine learning paradigm that is applied in hybrid intelligence systems can be categorized into four relevant subfields: *supervised, unsupervised, semi-supervised*, and *reinforcement* learning [52]. In *supervised learning*, the goal is to learn a function that maps the input data x to a certain output data y, given a labeled set of input-output pairs. In *unsupervised learning*, such output y does not exist and the learner tries to identify pattern in the input data x [16]. Further forms of learning such as reinforcement learning or semi-supervised learning can be subsumed under those two paradigms. *Semi-supervised learning* describes a combination of both paradigms, which uses both a small set of labeled and a large set of unlabeled data to solve a certain task [53]. Finally, *reinforcement learning*. An agent interacts with an environment thereby learning to solve a problem through receiving rewards and punishment for a certain action [5, 6].

**Human Learning Paradigm:** Humans have a mental model of their environment, which gets updated through events. This update is done by finding an explanation for the event [50, 49, 10]. Human learning can therefore can be achieved from *experience* and comparison with previous experiences [54, 44] and from description and *explanations* [55].

## 4.3. Human-AI Interaction

**Machine Teaching:** defines how humans provide input. First, humans can demonstrate actions that the machine learns to imitate [40]. Second, humans can annotate data for training a model for instance through crowdsourcing [56, 57]. We designate that as a *labeling*. Third, human intelligence can be used to actively identify a misspecification of the learner and debug the model, which we define as *troubleshooting* [58, 24]. Moreover, human teaching can take the form of *verification* whereby humans verify or falsify machine output [59].

**Teaching Interaction:** The input provided through human teaching, can be both *explicit* and *implicit*. While *explicit* teaching leverages active input of the user such as for instance labeling tasks such as image or text annotation [60], *implicit* teaching learns from observing the actions of the user and thus adapts to their demands. For instance, Microsoft uses contextual bandit algorithms to suggest users certain content, using the actions of the user as implicit teaching interaction.

**Expertise Requirements:** Hybrid intelligence systems can have certain requirements for the expertise of humans that provides input for systems. While by now both most research and practical applications focus on human input from an *ML expert* [61, 62, 63, 24], thus, requiring deep expertise in the field of AI. Moreover, *end users* can provide the system with input for product recommendations and e-commerce or input from human non-experts accessed through crowd work platforms [64, 58, 65]. More recent endeavors, however, focus on the integration of *domain experts* in hybrid intelligence architectures that leverage the profound understanding of the semantics of a problem domain to teach a machine, while not requiring any ML expertise [22, 66, 67].

**Amount of Human Input:** The amount of human input can vary between those of individual humans and aggregated input from several humans. *Individual* human input is for instance applied in recommender systems for individualization or due to cost efficiency reasons [60]. On the other hand, *collective* human input combines the input of several individual humans by leveraging mechanisms of human computation (e.g. [68, 66, 67]). This approach allows to reduce errors and biases of individual humans and the aggregation of heterogeneous knowledge [45, 69, 65].

**Aggregation:** When human input is aggregated from a collective of individual humans, different aggregation mechanisms can be leveraged to maximize the quality of teaching. First, *unweighted* methods can be used that use averaging or majority voting to aggregate results (e.g. [60]). Additionally, aggregation can be achieved by modeling the context of teaching through algorithmic approach such as expectation maximization, graphical models, entropy minimization, or discriminative training. Therefore, the aggregation can be *human dependent* focusing on the characteristics of a the individual human [70, 71, 72], or *human-task dependent* adjusting to the teaching task [73, 72, 74].

**Incentives:** Humans need to be incentivized to provide input in hybrid intelligence systems. Incentives can be *monetary rewards* such in the case of crowd work on platforms (e.g. Amazon Mechanical Turk), *intrinsic rewards* such as intellectual exchange in citizen science [75], fun in games with a purpose [76] learning [77]. Another incentive for human input is *customization*, which allows to increase individualized service quality for users that provide a higher amount of input to the learner [12, 78].

### 4.4. AI-Human Interaction

This sub-dimension describes the machine part of the Interaction, the AI-human interaction. At first, which query strategy the algorithm used to learn. Second, we describe the feedback of the machine to humans. Third, we carry out a short explanation of interpretability to show the influence for hybrid intelligence.

**Query Strategy:** *Offline* query strategies require the human to finish her task completely before her actions are applied as input to the AI (e.g. [79, 80]). Handling a typical labeling task the human would first need to go through all the data and label each instance. Afterwards the labeled Data is fed to an machine learning algorithm to train a model. In contrast, *online* query strategies let the human complete subtasks whose are directly fed to an algorithm, so that teaching and learning can be executed almost simultaneously [64, 58, 72]. Another possibility is the use of *active learning* query strategies [81, 23]. In this case, the human is queried by the machine when more input to give an accurate prediction is required.

**Machine Feedback:** Those four categories describe the feedback that humans receive from the machine. First, humans can get direct *suggestions* from the machine, which makes explicit recommendations to the user on how to act. For instance recommender systems such as Netflix or Spotify provide such suggestions for users. Furthermore, systems can make suggestions for describing images [58]. *Predictions* as machine feedback can support humans e.g. to detect lies [45], predict worker behaviors [72], or classify images. Thereby, this form of feedback provides a probabilistic value of a certain outcome (e.g. probability of some data x belonging to a certain class y). The third form of machine feedback is *clustering* data. Thereby, machines compare data points and put them in an order for instance to prioritize items [82], or organize data among identified pattern. Furthermore, another possibility of machine feedback is *optimization*. Machines enhance humans for instance in making more consistent decisions by optimizing their strategy [83].

**Interpretability**: For AI-Human interaction in ***hybrid intelligence*** systems interpretability is crucial to prevent biases (e.g. racism), achieve reliability and robustness, ensure causality of the learning, debugging the learner if necessary and for creating trust especially in the context of AI safety [11]. Interpretability in ***hybrid intelligence*** systems can be achieved through *algorithm transparency*, that allows to open the black box of an algorithm itself, *global model interpretability* that focuses on the general interpretability of a machine learning model, and *local prediction interpretability* that tries to make more complex models interpretable for a single prediction [84, 11].

## 5. Discussion

Our proposed taxonomy for ***hybrid intelligence*** systems extracts interdisciplinary knowledge on human-in-the-loop mechanisms in ML and proposes initial descriptive design knowledge for the development of such systems that might guide developers. Our findings reveal the manifold applications, mechanisms, and benefits of hybrid systems that might probably become of increasing interest in real-world applications in the future. In particular, our taxonomy of design knowledge offers insights on how to leverage the advantages of combining human and machine intelligence. For instance, this allows to integrate deep domain insights into machine learning algorithms, continuously adapt a learner to dynamic problems, and enhance trust through interpretability and human control. Vice versa, this approach offers the advantage of improving humans in solving problems by offering feedback on how the task was conducted or the performance of a human during that task and machine feedback to augment human intelligence. Moreover, we assume that the design of such systems might allow to move beyond sole efficiency of solving tasks to combined socio-technical ensembles that can achieve superior results that could no man or machine have achieved so far. Promising fields for such systems are in the field of medicine, science, innovation and creativity.

# 6. Conclusion

Within this paper we propose a taxonomy for design knowledge for *hybrid intelligence* systems, which presents descriptive knowledge structured along the four meta-dimensions *task characteristics, learning paradigm, human-AI interaction*, and *AI-human interaction*. Moreover, we identified 16 sub-dimensions and a total of 50 categories for the proposed taxonomy. By following a taxonomy development methodology [13], we extracted interdisciplinary knowledge on human-in-the-loop approaches in machine learning and the interaction between human and AI. We extended those findings with an examination of seven empirical applications of *hybrid intelligence* systems.

Therefore, our contribution is threefold. First, the proposed taxonomy provides a structured overview of interdisciplinary research on the role of humans in the machine learning pipeline by reviewing interdisciplinary research and extract relevant knowledge for system design. Second, we offer an initial conceptualization of the term *hybrid intelligence* systems and relevant dimensions for developing applications. Third, we intend to provide useful guidance for system developers during the implementation of hybrid intelligence systems in real-world applications.

Obviously this paper is not without limitations and provides a first step towards a comprehensive taxonomy of design knowledge on *hybrid intelligence* systems. First, further research should extend the scope of this research to more practical applications in various domains. By now our empirical case selection is slightly biased on decision problem contexts.Second, as we proceed our research we will further condensate the identified characteristics by aggregating potentially overlapping dimensions in subsequent iterations. Third, our results are overly descriptive so far. As we proceed our research we will therefore focus on providing prescriptive knowledge on what characteristics to choose in a certain situation and thereby propose more specific guidance for developers of *hybrid intelligence* systems that combine human and machine intelligence to achieve superior goals and driving the future progress of AI. For this purpose, we will identify interdependencies between dimensions and sub-dimensions and evaluate the usefulness of our artifact for designing real-world applications. Finally, further research might focus on integrating the overly design oriented knowledge of this study with research on knowledge-base systems in IS to discuss the findings in the context of those class of systems.